\documentclass[10pt,twocolumn,letterpaper]{article}

\usepackage{cvpr}
\usepackage{times}
\usepackage{epsfig}
\usepackage{graphicx}
\usepackage{amsmath}
\usepackage{amssymb}

\usepackage{color}

\usepackage{hyperref}
\usepackage{graphicx}
\usepackage[export]{adjustbox}

\usepackage{amsfonts}
\usepackage{booktabs}
\usepackage{siunitx}
\usepackage{pifont}
\usepackage{subcaption}
\newcommand{\cmark}{\ding{51}}%
\newcommand{\xmark}{\ding{55}}%
\usepackage{diagbox}
\usepackage{multirow}
\usepackage{pifont}

\cvprfinalcopy 

\def\cvprPaperID{4570} 

\newcommand*{\dprime}{^{\prime\prime}\mkern-1.2mu}
\ifcvprfinal\fi
\begin{document}

\title{Transferring Cross-domain Knowledge for Video Sign Language Recognition}

\author{Dongxu Li$^{1,2}$, Xin Yu$^{1,2,3}$, Chenchen Xu$^{1,4}$, Lars Petersson$^{1,4}$, Hongdong Li$^{1,2}$\\
$^{1}$The Australian National University, $^{2}$Australian Centre for Robotic Vision (ACRV),\\ $^{3}$University of Technology Sydney, $^{4}$DATA61-CSIRO\\
{\tt\small firstname.lastname@anu.edu.au}
}


\newcommand{\argmin}{\operatornamewithlimits{argmin}}
\newcommand{\argmax}{\operatornamewithlimits{argmax}}

\maketitle
\begin{abstract}
   Word-level sign language recognition (WSLR) is a fundamental task in sign language interpretation. It requires models to recognize \emph{isolated} sign words from videos. However, annotating WSLR data needs expert knowledge, thus limiting WSLR dataset acquisition. On the contrary, there are abundant subtitled sign news videos on the internet. Since these videos have no word-level annotation and exhibit a large domain gap from isolated signs, they cannot be directly used for training WSLR models.
   
   We observe that despite the existence of large domain gaps, isolated and news signs share the same visual concepts, such as hand gestures and body movements. Motivated by this observation, we propose a novel method that learns domain-invariant descriptors and fertilizes WSLR models by transferring knowledge of subtitled news sign to them. To this end, we extract news signs using a base WSLR model, and then design a classifier jointly trained on news and isolated signs to coarsely align these two domains. In order to learn domain-invariant features within each class and suppress domain-specific features, our method further resorts to an external memory to store the class centroids of the aligned news signs. We then design a temporal attention based on the learnt descriptor to improve recognition performance. Experimental results on standard WSLR datasets show that our method outperforms previous state-of-the-art methods significantly. We also demonstrate the effectiveness of our method on automatically localizing signs from sign news, achieving 28.1 for AP@0.5.

\end{abstract}
\section{Introduction}

Word-level sign language recognition (WSLR), as a fundamental sign language interpretation task, aims to overcome the communication barrier for deaf people. However, WSLR is very challenging because it consists of complex and fine-grained hand gestures in quick motion, body movements and facial expressions. 


%


\begin{figure}
    \centering 
\includegraphics[width=0.48\textwidth]{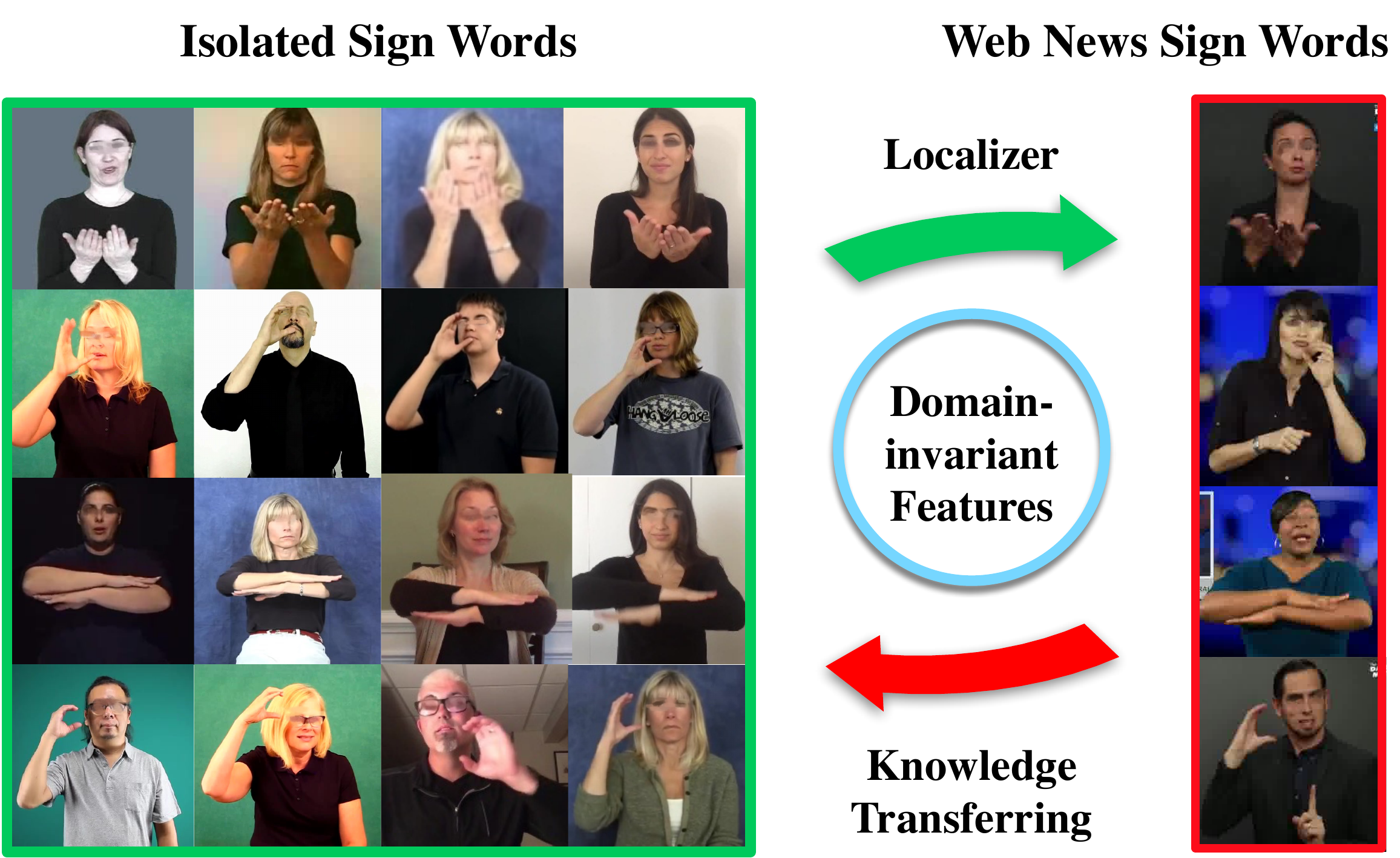}
    \vspace{-1.2em}
    \caption{Our model transfers the knowledge from web news signs to WSLR models by learning domain-invariant features. Example frames in the figure are identified by our model as the signature that best summarizes the gesture. }
    \label{fig:iconic}
    \vspace{-1em}
\end{figure}

Recently, deep learning techniques have demonstrated their advantages on the WSLR task~\cite{li2020word,joze2018ms,Pigou_2017_ICCV,kishore2018selfie}. However, annotating WSLR datasets requires domain-specific knowledge, therefore even the largest existing datasets have a limited number of instances, \eg, on average 10 to 50 instances per word~\cite{li2020word,joze2018ms,martinez2002purdue}. This is in an order of magnitude fewer than common video datasets on other tasks~\eg \cite{carreira_CVPR_2017_i3d}. The limited amount of training data for the sign recognition task may lead to overfitting or otherwise restrict the performance of WSLR models in real-world scenarios. %
On the other hand, there are abundant subtitled sign news videos easily attainable from the web which may potentially be beneficial for WSLR. 





Despite the availability of sign news videos, transferring such knowledge to WSLR is very challenging. First, subtitles only provide weak labels for the occurrence of signs and there is no annotation of temporal location or categories. Second, such labels are noisy. For example, a subtitle word does not necessarily indicate if the word is signed.
Third, news signs typically span over 9-16 frames~\cite{buehler2009learning}, which is significantly different from the videos (on average 60 frames~\cite{li2020word,joze2018ms}) used to train WSLR models in terms of gesture speed.
Therefore, directly augmenting WSLR datasets with news sign examples fails to improve recognition performance.



%

In this paper, we present a method that transfers the cross-domain knowledge in news signs to improve the performance of WSLR models. 
More specifically, we first develop a sign word localizer to extract sign words by employing a base WSLR model in a sliding window manner. Then, we propose to coarsely align two domains by jointly training a classifier using news signs and isolated signs. 
After obtaining the coarsely-aligned news words representations, we compute and store the centroid of each class of the coarsely-aligned new words in an external memory, called~\emph{prototypical memory}. 

Since the shared visual concepts between these domains are important for recognizing signs, we exploit prototypical memory to learn such domain-invariant descriptors by comparing the prototypes with isolated signs. In particular, given an isolated sign, we first measure the correlations between the isolated sign and news signs and then combine the similar features in prototypical memory to learn a \emph{domain-invariant descriptor}. In this way, we acquire representations of shared visual concepts across domains.


After obtaining the domain-invariant descriptor, we propose a \emph{memory-augmented temporal attention} module that encourages models to focus on distinguishable visual concepts among different signs while suppressing common gestures, such as demonstrating gestures (raising and putting down hands) in the isolated sign videos. Therefore, our network focuses on the visual concepts shared within each class and ignore those commonly appearing in different classes, thus achieving better classification performance.
%

In summary, (i)~we propose a coarse domain alignment approach by jointly training a classifier on news signs and isolated signs to reduce their domain gap;
(ii)~we develop prototypical memory and learn a domain-invariant descriptor for each isolated sign;
(iii)~we design a memory-augmented temporal attention over the representation of isolated signs and guide the model to focus on learning features from common visual concepts within each class while suppressing distracting ones, thus facilitating classifier learning;
%
(iv)~experimental results demonstrate that our approach significantly outperforms state-of-the-art WSLR methods on the recognition accuracy by a large margin of 12\% on WLASL and 6\% on MSASL. Furthermore, we demonstrate the effectiveness of our method on localizing sign words from sentences automatically, achieving 28.1 AP@0.5. 
Therefore, our method has a prominent potential for this process.
\section{Related Works}
Our work can be viewed as a semi-supervised learning method from weakly- and noisy-labelled data. In this section, we briefly review works in the relevant fields.


\subsection{Word-level Sign Language Recognition}
Earlier WSLR models rely on hand-crafted features~\cite{yasir2015sift,tharwat2015sift,yang2010chinese,benitez2014discriminant,ding2009modelling,cooper2012sign}. Temporal dependencies are modelled using HMM~\cite{starner1995visual,starner1998real} or DTW~\cite{lichtenauer2008sign}. Deep models learn spatial representations using 2D convolutional networks and model temporal dependencies using recurrent neural networks~\cite{li2020word,joze2018ms}. 
Some methods also employ 3D convolutional networks to capture spatio-temporal features simultaneously~\cite{huang2015sign,ye2018recognizing,li2020word,joze2018ms}. In addition, several works~\cite{ko2018sign,ko2019neural} exploit human body keypoints as inputs to recurrent nets. It is well known that training deep models require a large amount of training data. However, annotating WSLR samples requires expert knowledge, and existing WSLR video datasets~\cite{joze2018ms,li2020word} only contain a small number of examples, which limits the recognition accuracy. Our method aims at tackling this data insufficiency issue and improving WSLR models by collecting low-cost data from the internet.


\subsection{Semi-supervised Learning from Web Videos}
Some works~\cite{liang2016learning,yeung2017learning,ghadiyaram2019large} attempt to learn visual representations through easily-accessible web data. 
In particular, \cite{liang2016learning} combines curriculum learning~\cite{bengio2009curriculum} and self-pace learning~\cite{kumar2010self} to learn a concept detector. \cite{yeung2017learning} introduces a Q-learning based model to select and label web videos, and then directly use the selected data for training. Recently,~\cite{ghadiyaram2019large} found that pretraining on million-scale web data improves the performance of video action recognition. These works demonstrate the usefulness of web videos in a semi-supervised setting. 
Note that, the collected videos are regarded as individual samples in prior works. However, our collected news videos often contain multiple signs in a video, which brings more challenges to our task.

\begin{figure}[tb!]
\minipage{0.47\linewidth}
  \includegraphics[width=\linewidth]{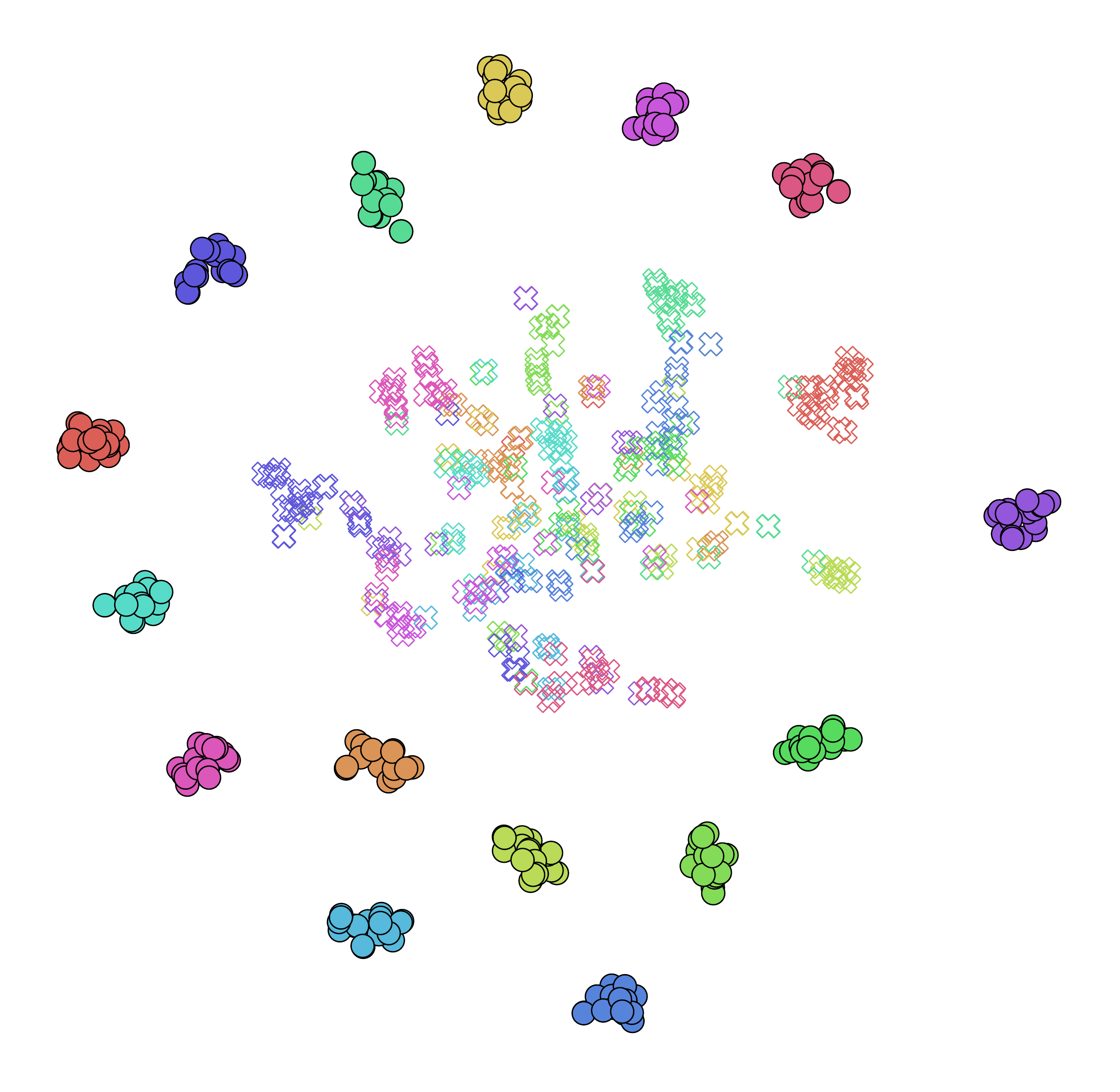}\subcaption{}\label{fig:tsnea}
\endminipage\hfill
\minipage{0.47\linewidth}
  \includegraphics[width=\linewidth]{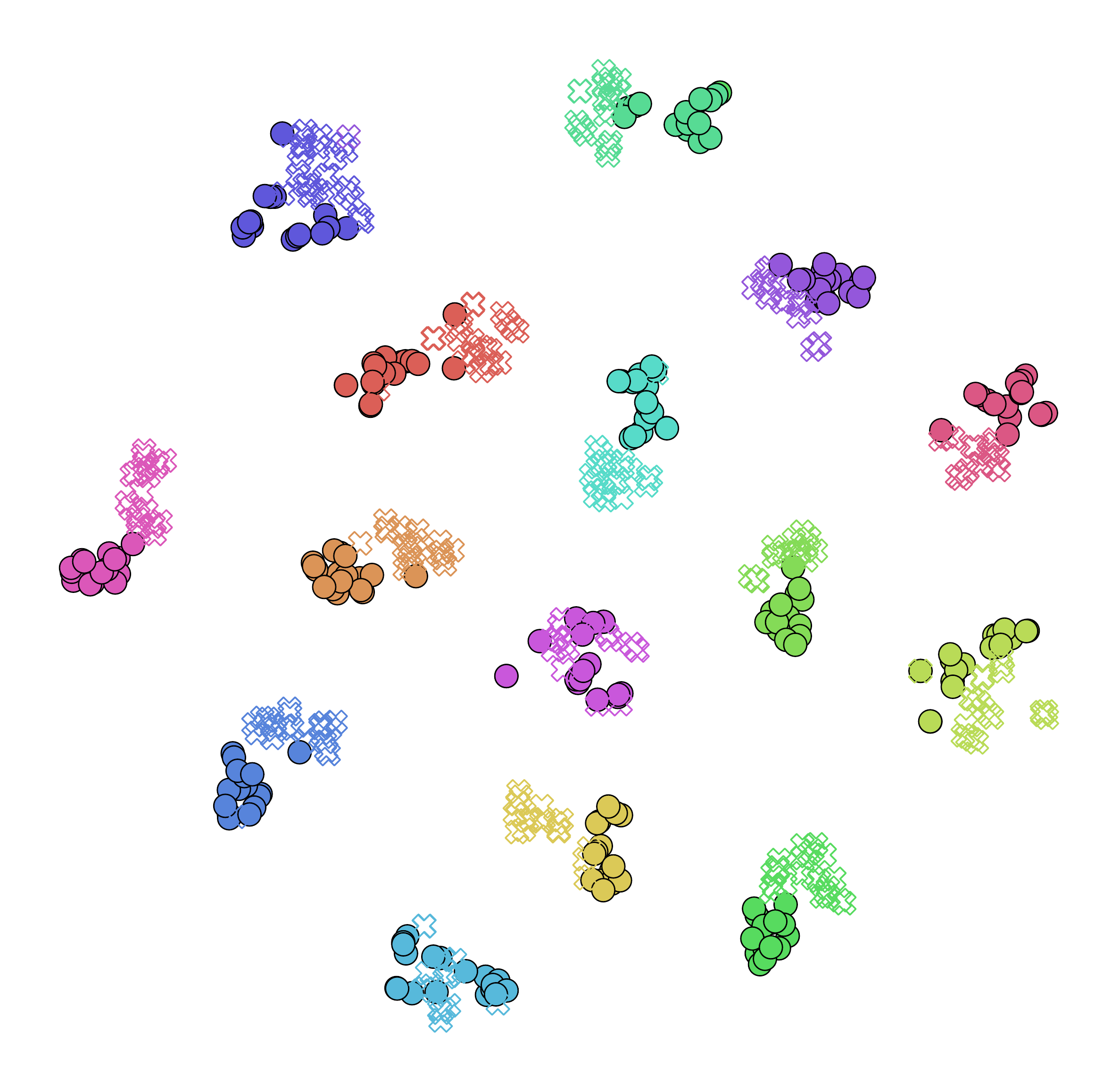}\subcaption{}\label{fig:tsneb}
\endminipage\hfill

\caption{Visualizing sign word training samples from two domains using t-SNE~\cite{maaten2008visualizing} before (a) and after (b) coarse domain alignment. Filled circles are isolated signs; empty crosses are news signs. Colors represent different classes.}
\label{img/preprocess}
\end{figure}
\subsection{Prototypical Networks and External Memory} 
Prototypical networks~\cite{snell2017prototypical} aim at learning classification models in a limited-data regime~\cite{zhang2019few}. During testing, prototypical networks calculate a distance measure between test data and prototypes, and predict using nearest-neighbour principle. A prototypical network provides a distance-based partition of the embedding space and facilitates the retrieval based on the nearest neighbouring prototypes in its essence.

External memory equips a deep neural network with capability of leveraging contextual information.
They are originally proposed for document-level question answering (QA) problems in natural language processing~\cite{weston2014memory,sukhbaatar2015end}.
Recently, external memory networks have been applied to visual tracking~\cite{na2017read}, image captioning~\cite{chunseong2017attend} and movie comprehension~\cite{yeung2017learning}. 
In general, external memory often serves as a source providing additional offline information to the model during training and testing.


\begin{figure*}[t!]
\centering
  \includegraphics[width=0.80\linewidth]{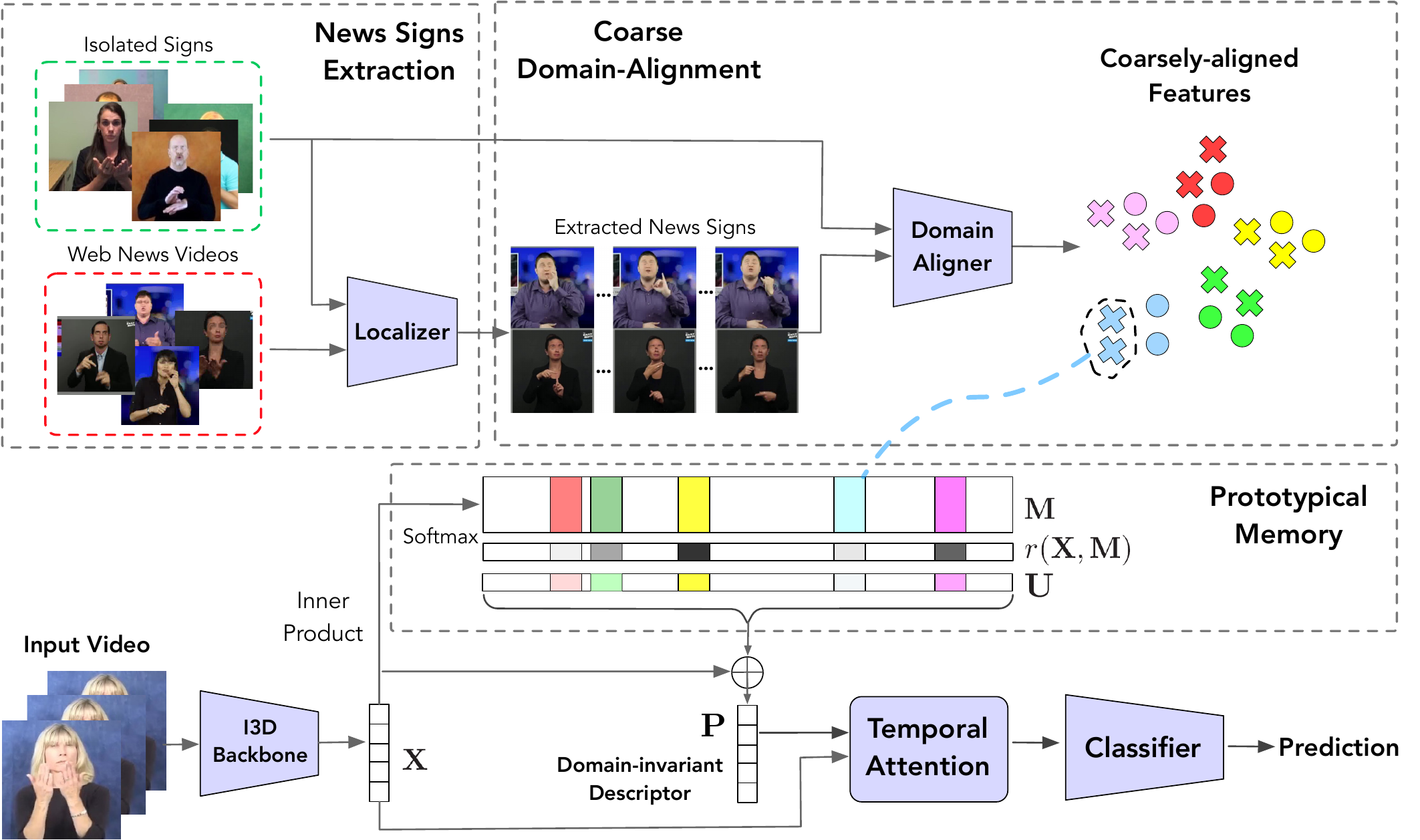}

\caption{Overview of our approach to transfer knowledge of coarsely aligned news signs to WSLR model using domain-invariant descriptor and memory-augmented temporal attention.}
\label{fig:overview}
\end{figure*}

\section{Proposed Approach}
\subsection{Notation}

A WSLR dataset with $N$ labeled training examples is denoted by $\mathcal{D}_s=\{I_i,\, L_i\}_{i=1}^{N}$, where $I_i\in \mathbb{R}^{l\times h\times w\times 3}$ is an input RGB video; $l$ is the number of frames (on average 64); $h$ and $w$ are the height and width of the frame respectively, and $L_i\in \mathbb{R}^{K}$ is a label of $K$ classes. We also consider a complementary set of sign news data denoted by $\mathcal{D}_n=\{S_i,\, T_i\}_{i=1}^{M}$. Similarly, $S_i$ is an RGB video, but with an average length of 300 frames. $T_i$ is a sequence of English tokens representing the subtitles corresponding to $S_i$.

\subsection{Overview}
We observe that despite the domain difference between signs from news broadcasts and isolated signs, samples from the same class share some common visual concepts, such as hand gestures and body movements. 
In other words, these shared visual concepts are more suitable to represent the cross-domain knowledge and invariant to domain differences.
Motivated by this intuition, we encourage models to learn such cross-domain features and exploit them to achieve better classification performance.

To this end, we first extract news signs from $S_i$ and train a classifier jointly using news and isolated signs.
In this fashion, we are able to coarsely align these two domains in the embedding space. Then, we exploit prototypes to represent the news signs and store in an external prototypical memory (Sec.~\ref{sec:proto-mem}). 
Furthermore, for each isolated sign video, we learn a domain-invariant descriptor from the external memory by measuring its correlation with the contents in each memory cell (Sec.~\ref{sec:proto-prior}). 
Based on our learnt domain-invariant descriptor, we design a memory-augmented temporal attention module to let isolated sign representation focus on temporally similar signing gestures, thus promoting the classification accuracy.
Figure \ref{fig:overview} illustrates an overview of our method.



\subsection{Constructing Prototypical Memory}\label{sec:proto-mem}
\subsubsection{Extracting words from weakly-labelled videos}
In order to utilize the data from news broadcasts, we need to localize and extract news signs from the subtitled videos. Specifically, we first pre-process the subtitles by lemmatizing~\cite{schutze2008introduction,li2018improving} the tokens and convert lemmas into lowercase. 
Then, for each isolated sign class $c_j, j= 1,..., K$, we collect video clips which contains the word $c_j$ in the processed subtitles. 
To do so, we apply a classifier $\mathcal{F}$ pretrained on isolated signs $\mathcal{D}_s$ to the collected videos in a sliding window manner. For each window, we acquire the classification score of each class $c_j$. 
For each video $S_i$, we choose the sliding window that achieves the highest classification score for $c_j$, \ie, $s^{*}_{ij}=\argmax_{s_i \sqsubset S_i}\mathcal{F}(c_j|s_{i})$, where $s_{i}\sqsubset S_i$ denotes that $s_i$ is a sliding window from $S_i$. 
Lastly, we discard windows with a class score lower than a threshold $\epsilon$. We use $S^*_j$ to denote the set of news sign video clips collected for $c_j$, \ie, $S^*_j = \{s^{*}_{ij}\mid\forall i:\: \mathcal{F}(c_j | s^{*}_{ij}) > \epsilon\}$.

\subsubsection{Joint training for coarse domain alignment}\label{sec:coarse}
Although $\mathcal{F}$ can exploit the knowledge learned from isolated signs to recognize news signs to some extent, we observe that $\mathcal{F}$ struggles to make confident predictions. 
In particular, $\mathcal{F}$ produces many false negatives and therefore misses valid news signs during the localization step. 
This is not surprising by acknowledging the domain gap. 
This phenomenon mainly comes from the domain gap between news signs and isolated ones.
As can be seen in Figure~\ref{fig:tsnea}, the features of isolated signs and news ones exhibit different distributions, which is undesirable when transferring knowledge between these two domains. 
To tackle this issue, we propose to first train a classifier jointly using sign samples from both domains, denoted by $\mathcal{\hat{F}}$. 

We use I3D~\cite{carreira2017quo,wang2019hallucinating} as the backbone network for both $\mathcal{F}$ and $\mathcal{\hat{F}}$. For feature extraction, we remove its classification head and use the pooled feature maps from the last inflated inception submodule. %
Figure~\ref{fig:tsneb} shows the feature representations of these two domain videos after the coarse domain alignment, where the domain gap is significantly reduced. 





\subsubsection{Prototypical memory}
In order to exploit the knowledge of news signs when classifying isolated signs, we adopt the idea of external memory.
We propose to encode the knowledge of news signs into \emph{prototypical memory}, where a \emph{prototype}~\cite{snell2017prototypical} is stored in a memory cell. Specifically, for class $c_j$, we define its prototype $\mathbf{m}_j$ as the mean of the feature embeddings of all the samples in $c_j$:
\begin{equation}
    \mathbf{m}_j = \frac{1}{|S_j^{*}|} \sum_{s_{ij}^{*} \in S_j^{*}}\mathcal{\hat{F}}(s_{ij}^{*}).
\end{equation}

A prototypical memory $\mathbf{M}\in \mathbb{R}^{K\times d}$ is constructed as an array of prototypes, i.e. $\mathbf{M} = [\mathbf{m}_1,\mathbf{m}_2,...,\mathbf{m}_K]$, where $d$ is the dimension of the prototype features.

Despite the abundance of sign news videos, the number of extracted samples is much less due to the domain gap.
Recall that our classifier $\mathcal{\hat{F}}$ is able to minimize the domain gap. It would be a solution of using $\mathcal{\hat{F}}$ to re-collect samples. However, we observe that the performance of the classifier $\mathcal{\hat{F}}$ on WSLR decreases and using $\mathcal{\hat{F}}$ to select news sign video clips does not generate more news sign samples.
This phenomenon can also be explained in Figure~\ref{fig:tsneb}. 
Since $\mathcal{\hat{F}}$ aims to minimize the domain gap, each cluster becomes less concentrated, which leads to the decrease of the classification accuracy.

Prototype representation provides us with a robust way to represent news signs in a limited-data regime. It induces a partition of the embedding space based on a given similarity measurement, which facilitates effective retrieval of similar visual concepts encoded in the news signs. 
By arranging them in an external memory, we link our classification model to a knowledge base of high-level visual features. In the next section, we will explain how to use these memory cells to learn a domain-invariant descriptor and then employ the domain-invariant feature to promote WSLR model.

\subsection{Learning Domain-invariant Descriptor}\label{sec:proto-prior}
After the two domains are coarsely aligned, our method will focus on learning domain-invariant descriptor using the prototypical memory. In this way, we are able to extract the common concepts from these two domains.
For a prototypical memory $\mathbf{M}\in\mathbb{R}^{K\times d}$ and an isolated sign feature $\mathbf{X} \in \mathbb{R}^{t\times d}$, where $t$ is determined by the number of the video frames, our goal is to generate a class-specific common feature from the prototypical memory. 

Since $\mathbf{x}_i$ and $\mathbf{m}_i$ are extracted by two different backbone networks $\mathcal{F}$ and $\mathcal{\hat{F}}$\footnote{For simplicity, we also refer the backbones of these two classifiers to as $\mathcal{F}$ and $\mathcal{\hat{F}}$}, these features are embedded in different spaces. Therefore, in order to measure the correlation between $\mathbf{X}$ and $\mathbf{M}$, we employ two different projection matrices to project these two space in to a common one first and then compute their normalized dot product in the common embedding space:
\begin{equation}\label{eq:corr}
    r(\mathbf{X}, \mathbf{M}) = \sigma\Big[\mathbf{X}\mathbf{W}_X(\mathbf{M}\mathbf{W}_M)^T\Big],
\end{equation}
where $\sigma(\cdot)$ is a softmax function, i.e., $\sigma(z)_i = {e^{z_i}}/{\sum_j e^{z_j}}$  applied in row-wise; $\mathbf{W}_X\in\mathbb{R}^{d\times d^{\prime}}$ and $\mathbf{W}_M\in\mathbb{R}^{d\times d^{\prime}}$ are two projection matrices for $\mathbf{X}$ and $\mathbf{M}$, respectively.

Eq.~\ref{eq:corr} defines the correlation between the isolated sign and the features in prototypical memory cells in the common embedding space. According to the feature correlations, we reweighted the features in the memory in the common embedding space, as follows:
\begin{equation}
\mathbf{U} = r(\mathbf{X},\mathbf{M}) \mathbf{M}(\mathbf{W}_M+\mathbf{W}_{\delta}),
\end{equation}
where the perturbation matrix $\mathbf{W}_{\delta}\in\mathbb{R}^{d\times d^{\prime}}$ allows our model to compensate for the errors during the domain alignment. We then map $\mathbf{U}$ back to the input space as a residual of $\mathbf{X}$ and finally acquire the domain-invariant descriptor $\mathbf{P}\in\mathbb{R}^{1\times d}$ via  maxpooling:
\begin{align}
    \mathbf{Z} &= \mathbf{U}\mathbf{W}_u + \mathbf{X},\\
    \mathbf{P} &= \text{maxpool}(\mathbf{Z}),
\end{align}
where $\mathbf{W}_u\in\mathbb{R}^{d^{\prime}\times d}$ is a linear mapping.  
Next, we explain how to utilize $\mathbf{P}$ to learn word sign representations.




\subsection{Memory-augmented Temporal Attention}\label{sec:attn}
Since collecting isolated signs from continuous sentences involves a laborious frame-by-frame annotation process, existing isolated sign datasets are mostly collected in controlled environments for demonstration purposes.
In particular, signs in isolated datasets often consist of demonstrating gestures, such as raising up or putting down the hands, and those gestures appear in sign videos regardless of words. This will increase the difficulty of learning a WSLR classifier since common gestures emerge in all the classes. A good WSLR model is supposed to focus on those discriminative temporal regions while suppressing demonstrating gestures.

%

Our attention module is designed to capture salient temporal information using the similarity between the domain-invariant descriptor $\mathbf{P}$ and the isolated sign representation $\mathbf{X}$. Since the domain-invariant descriptor $\mathbf{P}$ is acquired from the prototypical memory, we call our attention as memory-augmented temporal attention.
Specifically, because $\mathbf{P}$ and $\mathbf{X}$ represent different semantics and lie in their own feature space, we compute their similarity matrix $\mathbf{S}\in\mathbb{R}^{1\times t}$ by first projecting them into a shared common space:
\begin{equation}
    \mathbf{S} = \mathbf{P}\mathbf{W}_P(\mathbf{X}\mathbf{W}_Q)^T,
\end{equation}
where $\mathbf{W}_P$, $\mathbf{W}_Q$ are linear mappings in $\mathbb{R}^{d\times d{\dprime}}$. This operation compares the domain-invariant descriptor with the feature of an isolated sign on each temporal region in a pairwise manner. Then we normalize the similarity matrix $\mathbf{S}$ with a softmax function to create the attention map $\mathbf{A}\in\mathbb{R}^{1\times t}$:
\begin{equation}\label{eq:attnmap}
    \mathbf{A} = \sigma(\mathbf{S}).
\end{equation}
Eq.~\ref{eq:attnmap} indicates that the attention map $\mathbf{A}$ describes the similarity of $\mathbf{P}$ and $\mathbf{X}$ in the embedded common space. To acquire the attended features for isolated signs, we design a scheme similar to squeeze-and-excitation~\cite{hu2018squeeze}. In particular, we first introduce a linear mapping $\mathbf{W}_V\in\mathbb{R}^{d\times d{\dprime}}$ to embed $\mathbf{X}$ to a low-dimensional space for attention operation and then lift it up back to the input space of $\mathbf{X}$ using linear mapping $\mathbf{W}_O\in\mathbb{R}^{d{\dprime}\times d}$ with $d{\dprime}<d$.
%
%
Namely, our attended isolated sign representation $\mathbf{V}\in\mathbb{R}^{1\times d}$ is derived as follows:
\begin{equation}~\label{eq:squeeze}
    \mathbf{V} = \mathbf{A} (\mathbf{X}\mathbf{W}_V\mathbf{W}_O) 
\end{equation}

We remark Eq.~\ref{eq:squeeze} aggregates features along channels and therefore learns a channel-wise non-mutually-exclusive relationship, while~\cite{hu2018squeeze} aggregates feature maps across spatial dimension to produce descriptors for each channel. 
We then complement the feature representation of isolated signs with such channel-wise aggregated information by adding $\mathbf{V}$ as a residual to $\mathbf{P}$ for final classification.
In this way, our model learns to concentrate on features from salient temporal regions and explicitly minimizes the influence of irrelevant gestures.

\subsection{Optimization}
We adopt the binary cross-entropy loss function as in~\cite{carreira2017quo}. Specifically, 
%
%
given a probability distribution $p$ over different classes of signs, the loss $\mathcal{L}$ is computed as:
\begin{equation*}
    \mathcal{L} = -\frac{1}{NK}\sum_{i=1}^{N}\sum_{j=1}^{K}\Big[y_{ij}log(p_{ij}) + (1-y_{ij})log(1-p_{ij})\Big]
\end{equation*}
where $N$ is the number of samples in the batch; $K$ is the number of classes; $p_{ij}$ denotes the probability for the $i$-th sample belonging to the $j$-th class, and $y$ is the label of the sample. 


\section{Experiments}
\subsection{Setup and Implementation Details}\label{sec:setup}

\begin{table}[t]\caption{Statistics of datasets. We use \#class to denote the number of different classes in each dataset; train, validation, test denote numbers of video samples in each split.}
\vspace{-2mm}
\centering
\resizebox{\linewidth}{!}{\begin{tabular}{cccccc}
\toprule
\multicolumn{1}{l}{} & \multicolumn{2}{c}{\#class} & \multicolumn{1}{c}{Train} & \multicolumn{1}{c}{Validation} & Test \\ \midrule
WSASL100~\cite{li2020word}             & \multicolumn{2}{c}{100}     & 1442                      & 338                     & 258  \\
WSASL300~\cite{li2020word}             & \multicolumn{2}{c}{300}     & 3548                      & 901                     & 668  \\ \midrule
MSASL100~\cite{joze2018ms}             & \multicolumn{2}{c}{100}     & 3658 (-4\%)                    & 1021 (-14\%)                   & 749 (-1\%) \\
MSASL200~\cite{joze2018ms}             & \multicolumn{2}{c}{200}     & 6106 (-4\%)                     & 1743 (-15\%)              & 1346 (-1\%) \\\bottomrule
\end{tabular}
}
    \label{table:dataset}
\end{table}

\noindent{\textbf{Datasets.}} We evaluate our model on the WLASL~\cite{li2020word} and MSASL~\cite{joze2018ms} datasets. Both datasets are introduced recently supporting large-scale word-level sign language recognition. These videos record native American Sign Language (ASL) signers or interpreters, demonstrating how to sign a particular English word in ASL. 
%
%
We note that some links used to download MSASL videos have expired and the related videos are not accessible. As a result, we obtain 7\% less data for training (\cite{li2020word,joze2018ms} use both training and validation data for retraining models) and 1\% fewer videos for testing on MSASL. Therefore, results on MSASL should be taken as indicative. Detailed dataset statistics\footnote{MSASL misses partial data due to invalid download links as discussed in Section~\ref{sec:setup}. The percentage of missing data is shown in brackets.} are summarized in Table~\ref{table:dataset}. In all experiments we follow public dataset split released by the dataset authors.

\begin{figure*}[t!]
\centering
  \includegraphics[width=0.95\linewidth]{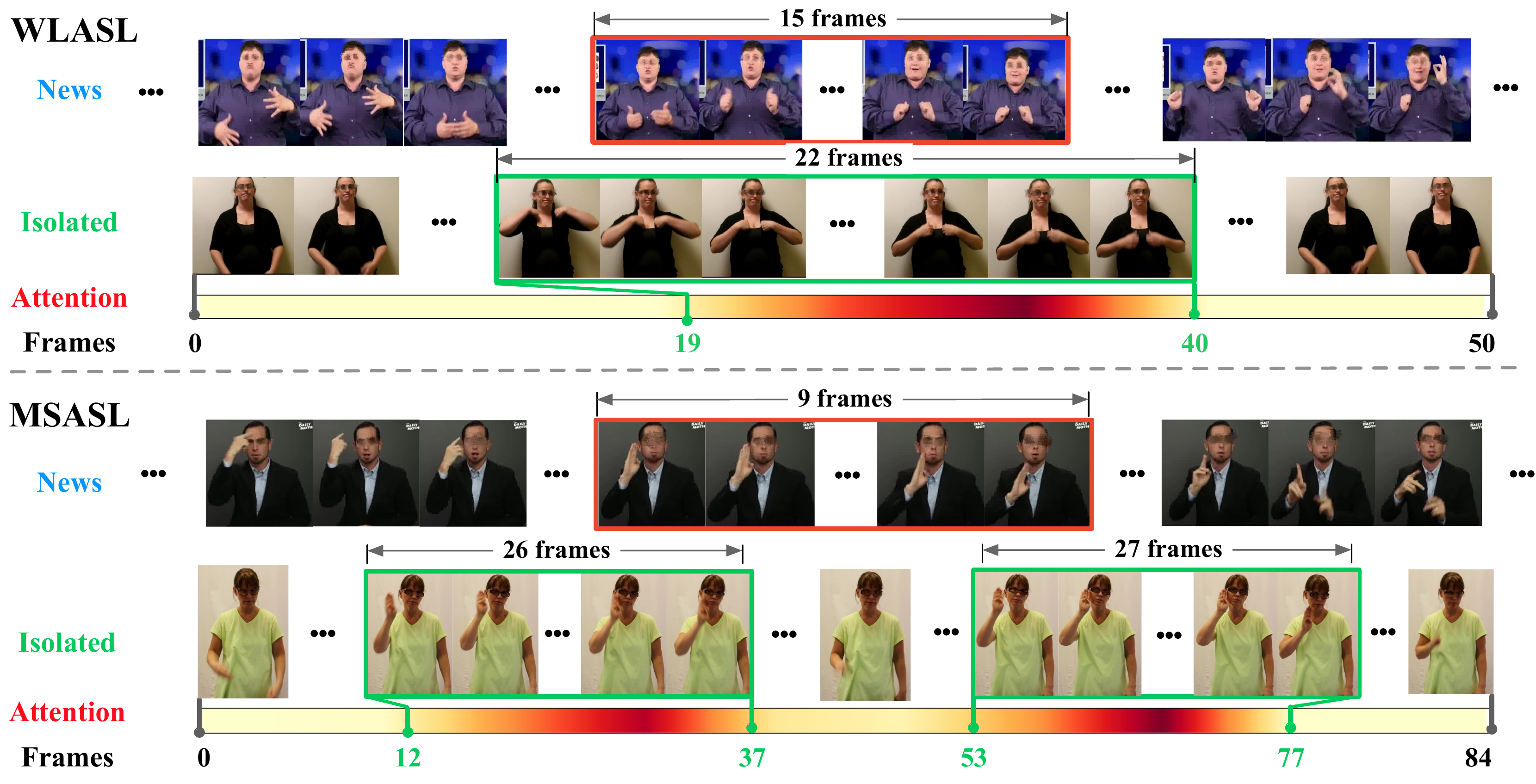}

\caption{Visualization of memory-augmented temporal attention on WLASL and MSASL. We present the extracted news signs in red boxes and isolated signs in green boxes. We use red to represent high-attention regions and light yellow for low-attention regions. }
\label{fig:attn}
\end{figure*}

\noindent{\textbf{Implementation details.}} Inflated 3D ConvNet (I3D)~\cite{carreira2017quo} is a 3D convolutional network originally proposed for action recognition. Considering its recent success on WSLR~\cite{li2020word,joze2018ms}, we use I3D as our backbone network and initialize it with the pretrained weights on Kinetics~\cite{carreira2017quo}. 
%
%
When extracting the word samples, we choose sliding windows of sizes 9$\sim$16 considering the common time span for a sign word~\cite{buehler2009learning}. We set threshold $\epsilon$ for localizing news signs to $0.3$ for WLASL100 and MSASL100, and to $0.2$ for WLASL300 and MSASL200, respectively.

\noindent{\textbf{Training and testing.}} We observe that although WLASL and MSASL datasets are collected from the different Internet sources, they have some videos in common. In order to avoid including testing videos in the training set, we do not merge the training videos from the two datasets. Instead, we train and test models on these two datasets separately.

Our training and testing strategies follow~\cite{li2020word,joze2018ms}. Specifically, during training, we apply both spatial and temporal augmentation. For spatial augmentations, we randomly crop a square patch from each frame. We also apply random horizontal flipping to videos because horizontally mirroring operation does not change the meaning of ASL signs. For temporal augmentation, we randomly choose 64 consecutive frames and pad shorter videos by repeating frames. 
We train our model using the Adam optimizer~\cite{kingma2014adam} with an initial learning rate of $10^{-3}$ and a weight decay of $10^{-7}$. During testing, we feed an entire video into the model. Similar to \cite{li2020word,joze2018ms}, we choose hyper-parameters on the training set, and report results by retraining on both training and validation sets using the optimal parameters. 

\subsection{Qualitative Results}

\noindent{\textbf{Visualizing memory-augmented attention}}
We visualize the output of the memory-augmented temporal attention in Fig.~\ref{fig:attn}. The first example is the word ``jacket" from WLASL. It can be seen that the temporal attention module filters out the starting and ending gestures in the video and learns to focus on the middle part of the video, where the sign is performed. The second example is the word ``brown" from MSASL. In this case, the attention map shows two peaks. By examining the video, we find that the sign is actually performed twice in a row with a slight pause in between. 
%

\noindent{\textbf{Generating sign signatures}}
The temporal attention facilitates to select representative frames from sign videos, referred to as ``sign signatures". 
In Fig.~\ref{fig:iconic}, the sign signatures are selected from the frames with the highest attention score from testing examples.
The sign signatures generated by our model are visually consistent with those manually identified from the news signs.
A potential usage for sign signatures is to help to automatically create summary , \eg, thumbnails, for videos on sign language tutorial websites.

\begin{table*}[ht!]\caption{Recognition accuracy (\%) on WLASL. RCNN refers to the Recurrent Convolution Neural Networks; I3D refers to the plain I3D setting; I3D + n.w. denotes the setting where extracted news words are directly added into the training set. We use macro. to denote the macro average accuracy and use micro. to denote the micro average accuracy. ($^*$) Results of MSASL are indicative due to the missing training data.}~\label{table:results}
\vspace{-2mm}
\centering
\resizebox{0.95\linewidth}{!}
{
\begin{tabular}{ccccccccc|cccccccc}
\toprule
               & \multicolumn{4}{c}{WLASL100} &  \multicolumn{4}{c|}{WLASL300} &  \multicolumn{4}{c}{MSASL100$^*$} &  \multicolumn{4}{c}{MSASL200$^*$} \\\cmidrule(r){2-5}
               \cmidrule(r){6-9}
               \cmidrule(r){10-13}
               \cmidrule(r){14-17}

               & \multicolumn{2}{c}{micro.}        & \multicolumn{2}{c}{macro.}       & \multicolumn{2}{c}{micro.}        & \multicolumn{2}{c|}{macro.}       & \multicolumn{2}{c}{micro.}        & \multicolumn{2}{c}{macro.}       & \multicolumn{2}{c}{micro.}        & \multicolumn{2}{c}{macro.}
               \\
               & top1 & top5 & top1 & top5 
               & top1 & top5 & top1 & top5 
               & top1 & top5 & top1 & top5 
               & top1 & top5 & top1 & top5 \\
               \midrule
RCNN~\cite{li2020word,joze2018ms} & 25.97 & 55.04 & 25.28  & 54.13 & 19.31 & 46.56 & 18.93 & 45.76 & 15.75 & 39.12 & 16.34 & 39.16 & 8.84 & 26.00 & 8.49 & 25.94\\
I3D~\cite{li2020word,joze2018ms}  & 65.89 & 84.11 & 67.01 & 84.58  &  56.14 & 79.94 & 56.24 & 78.38 & 80.91  & 93.46 & 81.94  & \textbf{94.13} & 74.29 & 90.12 & 75.32 & 90.80         \\\midrule
I3D + n.w.      &    61.63   & 82.56 & 62.18 & 82.72 & 54.19 & 80.69     & 54.71 & 80.99   & 77.70       &  \textbf{93.59}   & 75.41  & 90.34 & 75.40 & 90.34 & 76.68 & 90.69         \\
Ours      &    \textbf{77.52}   &  \textbf{91.08}    &  \textbf{77.55}  & \textbf{91.42} & \textbf{68.56} & \textbf{89.52} & \textbf{68.75} & \textbf{89.41} & \textbf{83.04} & 93.46 & \textbf{83.91} & 93.52 & \textbf{80.31} & \textbf{91.82} & \textbf{81.14} & \textbf{92.24} \\\bottomrule           
\end{tabular}
}
\end{table*}

\subsection{Baseline Models}
We compare with two baseline WSLR models, i.e., Recurrent Convolutional Neural Networks (RCNN) and I3D. Both RCNN and I3D are suggested in~\cite{li2020word,joze2018ms} to model the spatio-temporal information in word-level sign videos and achieve state-of-the-art results on both datasets. 
 
\noindent{\textbf{RCNN.}} In RCNN, it uses a 2D convolutional network to extract spatial features on frames. Then recurrent neural networks, such as GRU~\cite{chung2014empirical} or LSTM~\cite{hochreiter1997long}, are stacked on top of the convolutional network to model temporal dependencies. In our experiment, we use the implementation from~\cite{li2020word} which uses a two-layer GRU on top of VGG-16.

\noindent{\textbf{I3D.}} I3D~\cite{carreira2017quo} is a 3D convolutional neural network that inflates the convolution filters and pooling layers of 2D convolutional networks. I3D is recently adapted for WSLR~\cite{li2020word,joze2018ms} and achieves a prominent recognition accuracy. For WLASL, we use pretrained weights from the authors of~\cite{li2020word}. For MSASL, we report our reproduced results. 

\subsection{Quantitative Results}
\subsubsection{Comparison of Recognition Performance}
We report recognition performance on two metrics: (i) macro average accuracy (macro.), which measures the accuracy for each class independently and calculates the average, as reported in~\cite{joze2018ms}; (ii) micro average accuracy (micro.), which calculates the average per-instance accuracy, as reported in~\cite{li2020word}.  We summarize the results in Table~\ref{table:results}.

In Table~\ref{table:results}, I3D+n.w. results indicate that directly adding news signs to the training set does not help the training and even harms the model performance in most cases. This demonstrates the influence of the domain gap. Moreover, the degradation in performance also reveals the challenge of transferring knowledge from the news words to the WSLR models. We also notice that on MSASL200, the recognition accuracy improves after adding the news words despite the large domain gap. Although the improvement is minor, this shows the validity of our collected news sign videos.


As Table~\ref{table:results} shows, RCNN performs poorly mainly because its limited capacity to capture temporal motion dependency. Our proposed method surpasses previous state-of-the-art I3D model on both datasets. Because we use the same backbone network (I3D) as the baseline models, we conclude that the improvements come from the knowledge transferred from news words. Since the news words do not exhibit irrelevant artefacts such as idling and arm raising, they let the model focus more on the actual signing part in isolated words and produce more robust features. 


We observe that our proposed model outperforms previous state-of-the-art by a large margin on WLASL. This is because WLASL has even fewer examples (13-20 in each class) compared to MSASL (40-50). For fully supervised models, the number of examples in WLASL is very scarce and it requires an efficient way to learn good representations. In this regard, our proposed approach is able to transfer the knowledge from the news words and helps the learning process in such a limited-data learning regime.

\subsubsection{Word-level Classifier as Temporal Localizer}
Lack of training data is one of the main obstacles for both word-level and sentence-level sign language recognition tasks~\cite{bragg2019sign}. 
One such problem for sentence-level sign recognition is the lack of accurate temporal boundary annotations for signs, which can be useful for tasks such as continuous sign language recognition~\cite{koller2017re}. We employ our word-level classifier as a temporal localizer to provide automatic annotations for temporal boundaries of sign words in sentences.

\noindent{\textbf{Setup.}} Since there is no ASL dataset providing frame-level temporal annotations, we manually annotate temporal boundaries for 120 random news word instances to validate our ideas. The word classes are from WLASL100. 
Our expert annotators are provided with a news sentence and a isolated sign video. They are asked to identify the starting and end frame of the sign word in the news sentence.

\noindent{\textbf{Annotation quality control.}} We use temporal-IoU (tIoU) to verify the annotation quality, which is widely used to evaluate temporal action localization results~\cite{shou2016temporal}. For the two time intervals $I_1$ and $I_2$, their tIoU is computed as $\text{tIoU} = {(I_1 \cap I_2)}/{(I_1 \cup I_2)}$.
The initial average tIoU between the annotations is 0.73. We discard those entries with tIoU$<$0.5. For the remaining entries, an agreement is reached by discussion. We keep 102 annotated entries.

\noindent{\textbf{Results.}} We demonstrate the improvement of the word recognizer by localization accuracy. To this end, we employ classifiers in a sliding window fashion of 9-16 frames and identify a sign word if the predicted class probability is larger than 0.2.  We compare I3D with our model by computing mAP at different tIoUs. As shown in Table~\ref{table:tmap}, our method achieves higher localization performance. and provides an option for automatic temporal annotations. 

\begin{table}[t!]\caption{Comparison on temporal localization of sign words by mAP. Columns are different tIoU levels.}~\label{table:tmap}
\centering

\begin{tabular}{ccccc}
\toprule
tIoU     & 0.1 & 0.3 & 0.5 & 0.7 \\ \midrule
plain I3D~\cite{li2020word,joze2018ms}       &  27.4   &    23.9 &  15.3   &    02.4 \\
Ours &  \textbf{42.8}   &    \textbf{38.1} & \textbf{28.1}    &    \textbf{08.1}\\
\bottomrule
\end{tabular}
\vspace{-2mm}
\end{table}

\subsection{Analysis and Discussions}
We investigate the effect of different components of our model by conducting experiments on WLASL100.

\noindent\textbf{Effect of coarse domain alignment.} We first study the effect of coarse domain alignment as mentioned in Sec.~\ref{sec:coarse}. To this end, we extract features for news signs using classifier $\mathcal{F}$ without coarse alignment, and store class centroids as memory. In Table~\ref{table:eff-coarse}, the model achieves better performance when coarse alignment is used. By training $\mathcal{\hat{F}}$ jointly on samples from two domains, the classifier aligns the domains in the embedding space. And when coarse domain alignment is not applied, the domain gap leads to less relevant prototypes and prevents from learning good domain-invariant features.

\noindent\textbf{Effect of cross-domain knowledge.}
To investigate the influence of cross-domain knowledge, we explore three settings to produce the prototypical memory: (i) simulating the case where only isolated signs are available. As an alternative, we use $\mathcal{F}$ to extract features for isolated signs and use their class centroids as memory. In the remaining two settings, we investigate the effectiveness of news sign prototypes. To this end, we use $\mathcal{\hat{F}}$ to extract features for both isolated and news sign words: (ii) employing centroids of only isolated word features as memory; (iii) using both isolated and news word features to compute centroids. 

\begin{table}[t!]
\caption{Effect of coarse domain alignment on the recognition accuracy (\%). The ``wo. coarse align." row denotes the setting without coarse domain alignment. The ``w. coarse align." row shows results with coarse domain alignment.}\label{table:eff-coarse}
\centering

\begin{tabular}{ccccc}
\toprule
                  & \multicolumn{2}{c}{micro.} & \multicolumn{2}{c}{macro.} \\\cmidrule(r){2-3}
                  \cmidrule(r){4-5}
                  & top1         & top5        & top1         & top5        \\\midrule
wo. coarse align. &    70.93  &  87.21 & 71.30 &    86.25  \\
w. coarse align.  &     \textbf{77.52}   & \textbf{91.08}            &        \textbf{77.55}      &    \textbf{91.42}    
\\
\bottomrule
\vspace{0.5em}
\end{tabular}
\centering
\caption{Effect of sign news on the recognition accuracy (\%). Rows correspond to different settings to produce external memory. The ``model" column shows the model to extract features, with $\mathcal{F}$ the plain I3D and $\mathcal{\hat{F}}$ the I3D after coarse alignment. The ``memory" column indicates whether isolated signs (iso.) or news signs (news) are used.}
\label{tab:diff-mem}
\begin{tabular}{ccccccc}
\toprule
model                   & \multicolumn{2}{c}{memory} & \multicolumn{2}{c}{micro.}                            & \multicolumn{2}{c}{macro.}                            \\ \cmidrule(r){2-3}
\cmidrule(r){4-5}
        \cmidrule(r){6-7}
        & iso.  & news & top1 & top5 & top1 & top5\\\midrule
$\mathcal{F}$  &    \cmark    & \xmark & 72.48  & 89.92 & 72.80 & 89.80                     \\
$\mathcal{\hat{F}}$   &  \cmark &  \xmark             & 72.09                     & 87.21                     & 72.38                     & 86.75                     \\ 
$\mathcal{\hat{F}}$ &   \cmark &\cmark & 66.67 & 86.05 & 67.27 & 86.13 \\\midrule
$\mathcal{\hat{F}}$  & \xmark  & \cmark &  \textbf{77.52}   & \textbf{91.08}            &        \textbf{77.55}      &    \textbf{91.42}   
\\\bottomrule
\end{tabular}
\vspace{-0.5em}
\end{table}
As seen in Table~\ref{tab:diff-mem}, only using the aligned model with news signs as memory achieves best performance. We further analyze performance degradation in other settings as follows. Setting (i), the model only retrieves information from the isolated signs thus does not benefit from cross-domain knowledge. Setting (ii), the representations of isolated signs are compromised due to the coarse alignment, thus providing even worse centroids than (i). Setting (iii), averaging cross-domain samples produces noisy centroids since their embeddings are not well clustered. 

\section{Conclusion}
In this paper, we propose a new method to improve the performance of sign language recognition models by leveraging cross-domain knowledge in the subtitled sign news videos. We coarsely align isolated signs and news signs by joint training and propose to store class centroids in prototypical memory for online training and offline inference purpose. 
Our model then learns a domain-invariant descriptor for each isolated sign. 
%
Based on the domain-invariant descriptor, we employ temporal attention mechanism to emphasize class-specific features while suppressing those shared by different classes.
In this way, our classifier focuses on learning features from class-specific representation without being distracted.
Benefiting from our domain-invariant descriptor learning, our classifier not only outperforms the state-of-the-art but also can localize sign words from sentences automatically, significantly reducing the laborious labelling procedure.

\noindent\textbf{Acknowledgement.} HL's research is funded in part by the ARC Centre of Excellence for Robotics Vision (CE140100016),  ARC-Discovery (DP 190102261) and ARC-LIEF (190100080) grants, as well as a research grant from Baidu on autonomous driving.  The authors gratefully acknowledge the GPUs donated by NVIDIA Corporation.  We thank all anonymous reviewers and ACs for their constructive comments. 

{\small
\bibliographystyle{ieee_fullname}
\bibliography{egbib}
}

\end{document}